\documentclass[a4paper]{article}

\usepackage{INTERSPEECH2016}

\usepackage{graphicx}
\usepackage{amssymb,amsmath,bm}
\usepackage{textcomp}
\usepackage{cite}
\usepackage{amsmath}
\usepackage{tabularx}
\usepackage{hyperref}
\usepackage{multirow,array}

\ninept

\title{Neural Attention Models for Sequence Classification: \\Analysis and Application to  \\Key Term Extraction and Dialogue Act Detection}
\makeatletter
\def\name#1{\gdef\@name{#1\\}}
\makeatother \name{{\em Sheng-syun Shen \quad Hung-yi Lee}}

\address{Graduate Institute of Communication Engineering\\
National Taiwan University\\
{\footnotesize \tt r03942071@ntu.edu.tw, hungyilee@ntu.edu.tw}}

\begin{document}

  \maketitle
  \begin{abstract}
  %abstract 不 cite paper
 Recurrent neural network architectures combining with attention mechanism, or neural attention model, have shown promising performance recently for the tasks including speech recognition, image caption generation, visual question answering and machine translation. 
In this paper, neural attention model is applied on two sequence labeling tasks, dialogue act detection and key term extraction. 
In the sequence labeling tasks, the model input is a sequence, and the output is the label of the input sequence.
The major difficulty of sequence labeling is that when the input sequence is long, it can include many noisy or irrelevant part.
If the information in the whole sequence is treated equally, the noisy or irrelevant part may degrade the classification performance. 
The attention mechanism is helpful for sequence classification task because it is capable of highlighting important part among the entire sequence for the classification task.
The experimental results show that with the attention mechanism, discernible improvements were achieved in the sequence labeling task considered here. 
The roles of the attention mechanism in the tasks are further analyzed and visualized in this paper.  
%Recurrent neural network \cite{elman1990finding,jordan1997serial} architectures combining with attention mechanism have shown promising performance recently for some tasks, which includes speech recognition \cite{chorowski2015attention}, image caption generation \cite{xu2015show},  visual question answering \cite{xu2015ask}, and machine translation \cite{bahdanau2014neural}. This mechanism is capable of highlighting important part among the entire data, which may be helpful in some prediction tasks. We thus explore the role of attention mechanism in the tasks of language understanding. This paper focuses on the sub-problems of spoken language understanding, dialogue act classification and key term extraction, using a novel attention-based recurrent neural networks for solving problems. The experiments show encouraging performance on the AMI meeting corpus and a database collected from website articles.
  \end{abstract}
  \noindent{\bf Index Terms}: attention model, key term extraction, dialogue act detection, long short-term memory (LSTM)

%%%%%%%%%%%%%%%%%%%%%%%%%%%%%%%%%%%%%%%%%%%%%%%%%%%%%
%-------------------------------------------------------------------------------------------------------------------------------%
%%%%%%%%%%%%%%%%%%%%%%%%%%%%%%%%%%%%%%%%%%%%%%%%%%%%%

  \section{Introduction} \label{sec:intro}

Recently, attention-mechanism has been incorporated with recurrent neural networks, and has shown significant improvement on a great variety of tasks. 
Attention mechanism is first introduced by Bahdanau et al. \cite{bahdanau2014neural} in the task of machine translation. They proposed an recurrent neural network (RNN)\cite{elman1990finding,jordan1997serial} encoder-decoder model for end-to-end translation, and this mechanism is intuitively designed in order to take care about the positions of input elements according to previous output result. Inspired by this work, Chorowski et al. \cite{chorowski2015attention} then proposed attention-based models for speech recognition, which are claimed to be robust to long inputs. Kelvin Xu et al. \cite{xu2015show} and Huijuan Xu et al. \cite{xu2015ask} also demonstrated how attention mechanism works while reading a picture. The above works iteratively process their input by selecting relevant content at every step. Attention-mechanism are also useful for tasks other than sequence to sequence learning. Memory Neural Networks (MemNN) which are developed by Weston et al.~\cite{weston2014memory} and Sukhbaatar et al. ~\cite{sukhbaatar2015end} can deal with question answering (QA) task~\cite{1603.07044,weston2014memory,sukhbaatar2015end}, and the attention-mechanism plays an important role in the model.

In this paper, neural attention model is applied on sequence classification tasks.
In a sequence classification task, the input of the model is a sequence, and the model output is the class of the sequence.
Many common tasks can be formulated as sequence classification including speaker recognition~\cite{IvectorIS09}, audio emotion classification~\cite{EmotionChallengeIS09}, spoken term detection (STD)~\cite{MyJournal_SVM,segment2vectorIS13,segment2vectorIS12}, dialogue act detection~\cite{louwerse2006dialog,boyer2011affect,surendran2006dialog}, key term extraction~\cite{sarkar2010new,chen2010automatic,nakagawa2002simple,chen2013empirical}, etc.  
One of the major difficulties for sequence classification is that when the input sequence is long, it can include many noisy or irrelevant parts, and without techniques to ignore these parts, they may degrade the classification performance.
Attention-mechanism shows the potential of automatically ignoring the unimportant parts in the entire input sequence and highlighting the important parts~\cite{1603.07044,weston2014memory,sukhbaatar2015end}. 
This inspires us to explore the use of attention mechanism on sequence classification. 

In this paper, we present a novel attention-mechanism long short-term memory (LSTM) \cite{gers2001lstm,hochreiter1997long} network architecture for sequence classification, in which the LSTM network reads the entire input, attention-mechanism highlights the important elements, and the sequence classes are predicted by the highlighted parts.  
This model is first tested on dialogue act detection in which the model input is the transcriptions of one to several utterances, and the output is the dialogue acts.
It is shown that the attention-mechanism is especially helpful with longer input.  
We further formulate the key term extraction as sequence classification task~\cite{sarkar2010new}, and apply the proposed model. 
This methodology shows promising results on key term extraction. Finally, visualization and analysis are also performed to understand how the attention process works.

\section[Neural Attention Model for  Sequence Classification]{Neural Attention Model for\\  Sequence Classification}
\label{sec:propose}

\begin{figure}[t]
        \centering
        \includegraphics[width=1.0\linewidth]{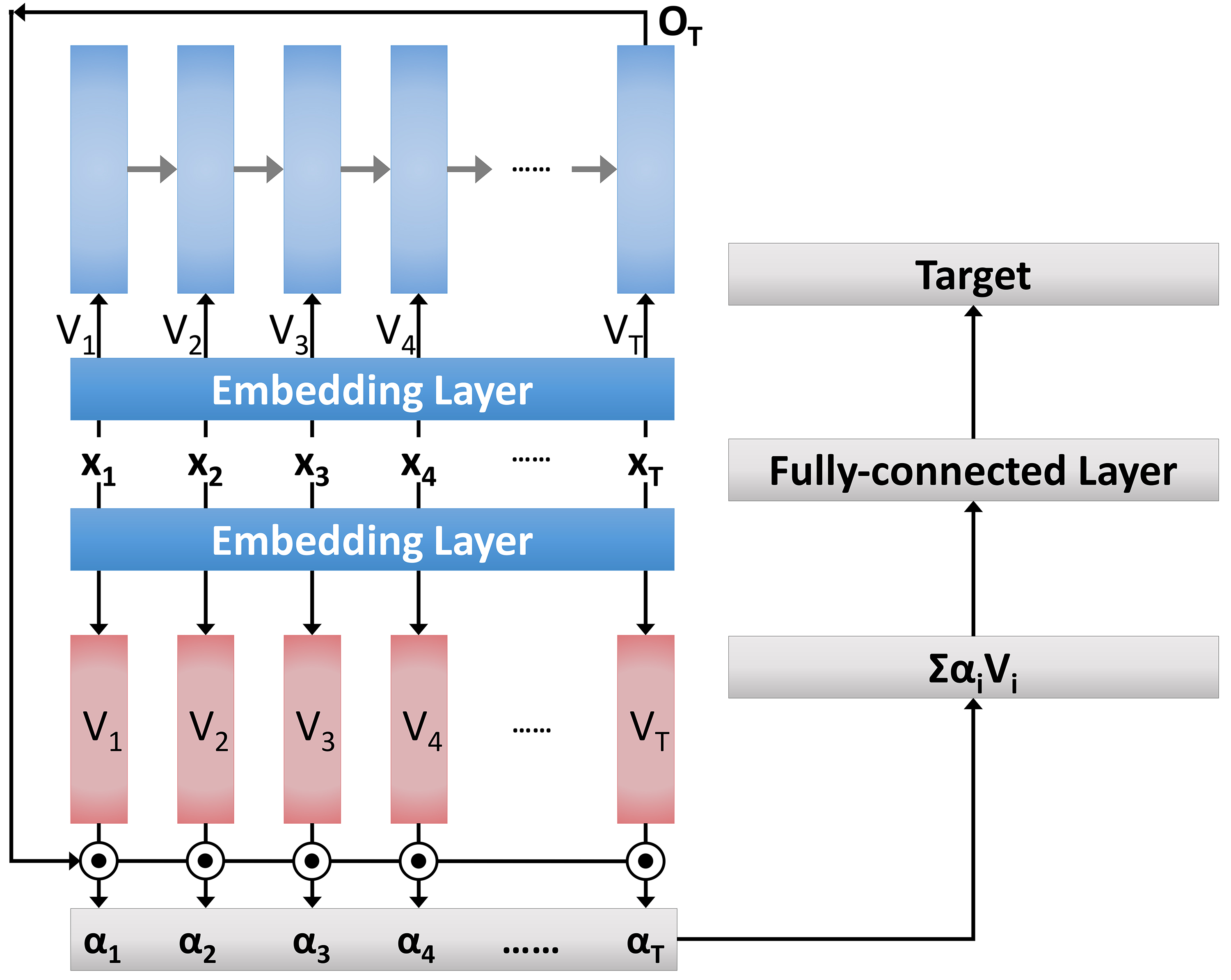}
		  \caption{{Architecture of the proposed Neural Attention Model. }}
        \label{fig:model}
      \end{figure}

The  overall structure of the proposed method is in Figure \ref{fig:model}. The inputs of model would be represented as a dense sequence vector $O_{T}$, which will be described in section \ref{subsec:seq}. With the sequence vector, attention mechanism is then applied to extract related information from input sequence in section \ref{subsec:att}. In section \ref{subsec:target}, the model will predict target according to the selected feature vectors.

  \subsection{Sequence Representation} \label{subsec:seq}
We use recurrent neural networks (RNN) for encoding. RNNs are capable of handling sequence information over time, so they have demonstrated outstanding performance on natural language understanding tasks \cite{yao2014spoken,yao2013recurrent,mesnil2015using} in recent years. 
  We  select long short-term memory (LSTM) networks, a type of recurrent neural networks with a more complex computational unit, to processes inputs sequentially. A brief introduction of LSTMs can be found in \cite{gers2001lstm,hochreiter1997long}.
  
  In the upper part of Figure \ref{fig:model}, we demonstrate the encoding procedure to transform input sequences into fixed-length vector representation $O_{T}$. The set $x = (x_{1}, x_{2}, \ldots , x_{T})$ denotes the input sequence, where $T$ is the sequence length. Each element in $x$ represents a fixed-length feature vector. For example, it might be a high dimensional 1-of-N encoding unigram vector for the task of text classification. 
  In order to reduce the model complexity, we set an embedding layer, a linear transformation matrix, to turn the inputs into low dimensional dense vectors $V=(V_{1}, V_{2}, \ldots, V_{T})$, and then they will be sent to the LSTM encoder.
  In each time step, the LSTM takes one element $V_{i}$ from feature vector set, and after processing the last element, it then generates an output vector $O_{T}$, which can be regarded as the summaries of the preceding feature vectors.

%The LSTM networks described by the following composition function:
%      \begin{alignat}{5}
%        i_{t} &= \sigma(W_{xi}x_{t} + W_{hi}h_{t-1} + W_{ci}c_{t-1}+b_{i}), \label{eq1} \\
%        f_{t} &= \sigma(W_{xf}x_{t} + W_{hf}h_{t-1} + W_{cf}c_{t-1}+b_{f}), \label{eq2}\\
%        c_{t} &= f_{t} \odot c_{t-1} + i_{t} \odot tanh(W_{xc}x_{t} + W_{hc}h_{t-1} + b_{c}), \label{eq3}\\
%        o_{t} &= \sigma(W_{xo}x_{t} + W_{ho}h_{t-1} + W_{co}c_{t}+b_{o}), \label{eq4}\\
%        h_{t} &= o_{t} \odot tanh(c_{t}), \label{eq5}
%       \end{alignat}
%  where $\sigma$ is the logistic sigmoid function. $i$, $f$, $o$ and $c$ are respectively for the input gate, forget gate, output gate, and memory cell activation vectors, all of which have the same size as the hidden vector $h$. The symbol $\odot$ denoted the element-wise product of the vectors. The weight matrices from the cell to gate vectors are diagonal, so element $m$ in each gate vector only receives input from element $m$ of the cell vector. 
%  The weight matrices from input, hidden, and outputs are not diagonal.

\subsection{Attention Mechanism} \label{subsec:att}
When input sequence $x$ is long, the summaries vector $O_{T}$ is likely to contain noisy information from many irrelevant feature vectors $V_{i}$, we thus apply attention mechanism to select only relevant frames among the entire sequence. The procedures are shown in the lower part of Figure \ref{fig:model}. There is also an embedding layer to transform input sequences into dense vectors, and all the parameters in the embedding layer are shared with the previous one. We then calculate the cosine similarity between the sequence vector $O_{T}$ and word embedding set $V$:
      \begin{equation} 
		e_{i} = O_{T}  \odot V_{i} , \label{cosine_sim}
\end{equation}
where $\odot$ denotes cosine similarity between two vectors. As a result, we have a list of score $e = (e_{1}, e_{2}, \ldots, e_{T})$. The attention weights $\alpha = (\alpha_{1}, \alpha_{2}, \ldots , \alpha_{T})$ come from the normalized score list $e$. Due to some considerations, we normalize the scores in two ways, which is inspired by Chorowski et al. in \cite{chorowski2015attention}:

  \textbf{Sharpening}: The score list is normalized using $softmax$ activation function:
     \begin{equation} 
%     \alpha_{i} = exp(e_{i}) \mathbin{/} \sum_{i=1}^{T} exp(e_{i}) , \label{sharp_norm}
		\alpha_{i} = \frac{exp(e_{i})}{\sum_{i=1}^{T} exp(e_{i})} , \label{sharp_norm}
\end{equation}
It has been widely used in many existing neural attention frameworks \cite{xu2015ask, xu2015show, bahdanau2014neural, weston2014memory, 1603.07044}, and is capable of solving the data noisy issue. 

  \textbf{Smoothing}: The sharpening normalization method prefers to mostly focus on only a single feature vector $V_{i}$, and might negatively affects the model's performance. We then apply a new way for the model to aggregate selections from multiple top-scored frames. In this way, more input locations are considered for bringing more diversity to the model. We replace the exponential function in equation (\ref{sharp_norm}) with logistic sigmoid function $\sigma$ :
       \begin{equation} 
%  		\alpha_{i} = \sigma(e_{i}) \mathbin{/} \sum_{i=1}^{T} \sigma(e_{i})  \label{smooth_norm}
		\alpha_{i} = \frac{\sigma(e_{i})}{\sum_{i=1}^{T} \sigma(e_{i})}  \label{smooth_norm}
\end{equation}
Visualization and analysis of the both normalization functions are provided in the experiment section.

  \subsection{Target Selection} \label{subsec:target}
The right part of Figure \ref{fig:model} illustrates the target selection procedures. We weighted sum all the feature vectors as $\sum \alpha_{i} V_{i}$, and sending it to a fully connected layer. Usually, the neurons in this layer are activated by nonlinear functions. The last layer is for target prediction, and the dimension is set to be candidate target numbers.

%%%%%%%%%%%%%%%%%%%%%%%%%%%%%%%%%%%%%%%%%%%%%%%%%%%%%
%-------------------------------------------------------------------------------------------------------------------------------%
%%%%%%%%%%%%%%%%%%%%%%%%%%%%%%%%%%%%%%%%%%%%%%%%%%%%%

  \section{Experiments} \label{sec:exp}
We conducted two sequence classification tasks in this section. In section \ref{subsec:DA}, we describe the definition of dialogue act detection, and also demonstrate the experimental results. 
In section \ref{subsec:KE}, we introduce how to apply the proposed methodology on key term extraction task. 
The role of attention mechanism during classification procedure will be discussed in section \ref{subsec:vis}, and we also show the visualization results.

  %These two tasks are important due to the development of World Wide Web (WWW). The huge amount of information over the Internet has made the need of retrieving, indexing, and categorizing these documents. With a view to achieve an efficient browsing and retrieval process for users, document categorization and index term extraction are effective ways for pre-processing procedures. Since there are different forms of documents on the Internet, we would like to implement dialogue act classification for multimedia data, and applying key term extraction for both spoken and written documents.

\subsection{Dialogue Act Detection} \label{subsec:DA}
Dialogue act (DA) detection \cite{louwerse2006dialog,boyer2011affect,surendran2006dialog} is about categorizing the intention behind the speaker's move in conversations, and recognition of a speaker's act may help reason the entire dialogue. 
This prediction task is still challenging because there are various distinct ways of formulating an intention. In this work, DAs are labeled with one of a number of tags. For example, the tag \texttt{<OFFER>} is related to the situation that someone commands partner to carry out actions, e.g., \textit{``You need to give me your ideas, and then I need to see whether that would sell in the market place.''}

\subsubsection{Experimental setup}
We conducted experiments on Switchboard Dialog Act (SwDA) Corpus \cite{jurafsky1997switchboard}, which is a corpus of telephone conversations on selected topics. It consists of about 2,500 conversations by 500 speakers from the U.S. The conversations in the corpus are labeled with 43 unique dialogue act tags and split to 1,115 train and 19 test conversations. The training and testing corpus respectively contain 213,543 and 4,514 utterances, having average length of about 8 words.

\subsubsection{Baselines} \label{subsec:baseline}
We compared the proposed model with the following baselines.

      \textbf{Support Vector Machines}: SVM is the most common way to be adopted for text classification. Silva et al. \cite{silva2011symbolic} chose sentence unigrams as input feature vector, and trained the SVM model. We extracted one-of-N encoding unigram features for every word in the dataset, aggregating them together for each training example. To reduce the number of dimensions, we set minimum word counts to 5. The Radial basis function (RBF) \cite{chang2010training} kernel was also applied.
      
        \textbf{Multiple Layer Perceptron}: The work introduced by Ries et. al \cite{ries1999hmm} is the first approach that importing artificial neural networks (ANN) for dialogue act detection. We also extracted unigram features as the model input for experiments. We trained an MLP model with 3 hidden layers. Each hidden layer has 512 neurons.
        %, and the structure detail was $512 \times 512 \times 512$. 
        The $relu$ activation function was applied on every hidden layer, and we set $rmsprop$ as the optimizer. The training epoch was set to be 20.
        
%shows the whole architecture of our MLP network. Each word in the input document would be transformed to a fix-length dense vector using some pre-trained word embedding models (e.g., Word2vec [x] and GloVe [x]) or a linear-projection matrix training with the entire MLP network jointly. We then sum up all the vectors, feeding this feature vector into hidden layers to predict results. Implementation details including the network structure and hyper-parameters setting would be briefly described in section \ref{sec:exp}.
        \textbf{Long Short-term Memory}: In order to examine the use of attention mechanism, we also implemented the original LSTM network. The LSTM model takes one word from the input sequence in each time step. We applied word embedding for unigram features, thus the high dimensional sparse vectors are transformed into dense vectors. The embedding size was 400, and we set the dimension of recurrent layers as 128 and the fully connected layer before output as 500, respectively. To avoid overfitting, we only trained the LSTM network for 10 epochs.

\subsubsection{Experimental results}
We implemented both sharpening-attend and smoothing-attend neural attention model in the experiments. 
The LSTM part of the proposed model is the same as the original LSTM briefly illustrated in the previous subsection, and the hyper-parameters for model training was also the same.
As the previous work stated \cite{ribeiro2015influence}, context information from previous utterances may help for the dialogue act prediction. Therefore, we also appended  $n$ previous utterances to the the utterance being classified, and $n$ was set to be 3 in the experiments.

The results are reported in Table \ref{tab:da_scor}. 
Rows (a) to (d) are the baseline results, and  the results of the proposed approaches are in rows (e) to (h). 
It is clear that the LSTM networks already outperformed the other baselines (rows (c) vs (a), (b)) because the LSTM networks have better capability of handling sequence information than multiple layer perceptrons and support vectors. 
Moreover, with context information the LSTM can have higher accuracy than the one without it (rows (d) vs (c)). 

Considering the case without context information, the proposed approaches show improvements comparing to all the baselines no matter the attention is sharpening or smoothing (rows (e), (f) vs (a), (b), (c)). 
The neural attention model with sharpening attention is only slightly better than the original LSTM (rows (e) v.s. (c)), but the smoothing attention shows significant improvement (rows (f) v.s. (c)). 
%The results also show that the smoothing normalization method can offer more information than sharpening normalization (rows (f) vs (e)). 
Besides, we also know that the prediction of sequence classification cannot just rely on the most relevant element, the rest of the relevant part should also be considered.
Neural attention model with sharpening attention does not show any improvement after adding context information into the prediction procedure (rows (g) v.s. (e)).
This is because the sharpening-attend mechanism only focuses on the most relevant part of the input sequence, adding more candidates would not be helpful.
On the other hand, when using smoothing attention, context information became very helpful (rows (h) v.s. (f)).
This shows that smoothing attention can better exploit the context information than sharpening attention.

%While adding context information into the prediction procedure, the smooth-attend model also outperformed the rest LSTM networks (rows (h) vs (d)(f)). 
%The interesting thing is despite the fact that adding context information into consideration, the sharp-attend LSTM still has the same accuracy in comparison with the one without consideration (rows (f) vs (e)). 

     \begin{table}[th]
        \caption{\label{tab:ke} {\it SwDA dialogue act detection accuracies.}}
        \label{tab:da_scor}
        \vspace{2mm}
        \centerline{ 
          \begin{tabular}{| l |l | c |}
            \hline
            \multicolumn{2}{| l |}{\textbf{Model}} & \multicolumn{1}{c|}{ \textbf{Accuracy (\%)}} \\ 
            \hline \hline
               \multicolumn{2}{|l|}{(a) Support Vector Machine} &       $65.8$  \\ \hline
               \multicolumn{2}{|l|}{(b) Multiple Layer Perceptron} &         $67.3$  \\  \hline
               \multicolumn{2}{|l|}{(c) Long Short-term Memory } &  $69.7$  \\ \hline
               \multicolumn{2}{|l|}{(d) LSTM with context information} &        $71.7$ \\ \hline                              
                 \multirow{2}{*}{Neural Attention Model}  & (e) Sharpening &       $69.9$ \\ \cline{2-3}
                 & (f) Smoothing  &        $70.4$ \\ \hline
     Neural Attention Model  & (g) Sharpening &       $69.8$ \\ \cline{2-3}
       with context information          & (h) Smoothing  &        $\textbf{72.6}$ \\ 
      %           \multirow{2}{*}{Sharp-Attend}  & (e) LSTM &       $69.9$ \\ \cline{2-3}
       %          & (f) LSTM with context  &        $69.8$ \\ \hline
        %         \multirow{2}{*}{Smooth-Attend}  & (g) LSTM &       $70.4$ \\ \cline{2-3}
         %        & (h) LSTM with context  &        $\textbf{72.6}$ \\ 
            \hline
          \end{tabular}
        }
      \end{table}
      
             \begin{figure*}[t]
        \centering
        \includegraphics[width=0.92\linewidth]{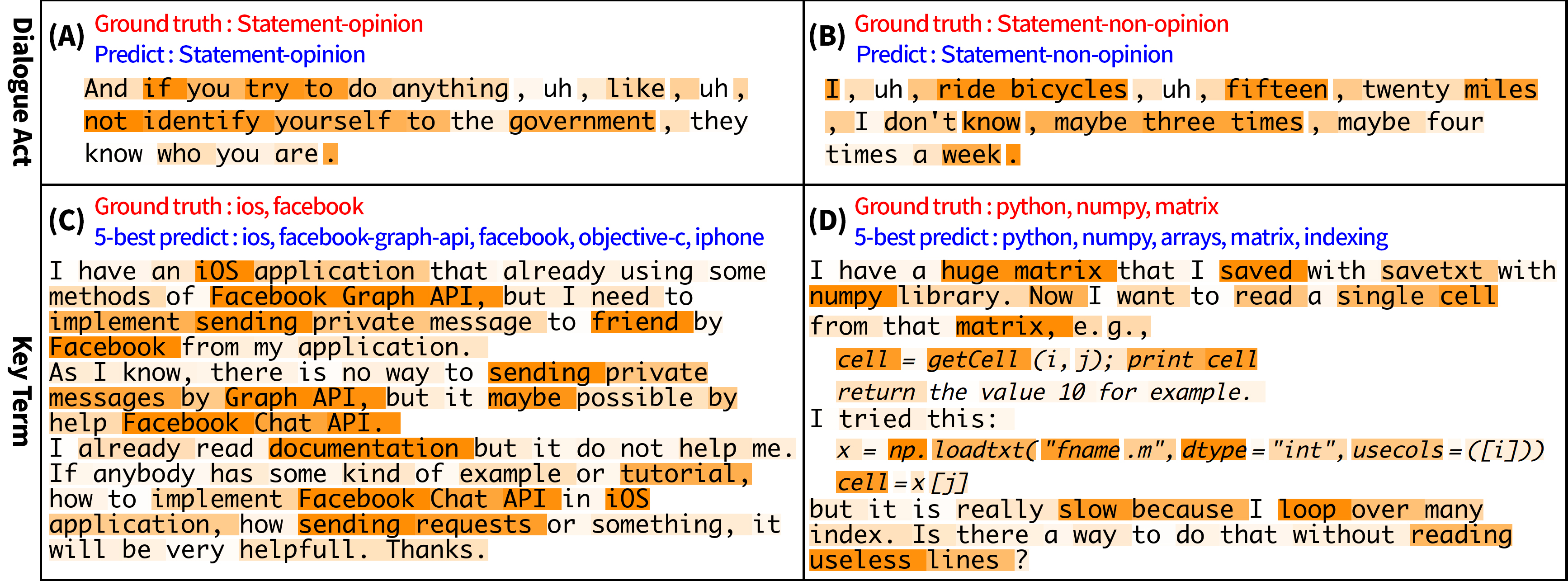}
        \caption{{\it Attention-mechanism visualization for sequence classification tasks. The darker color a word get, the higher attention weight it has. Figures (A) and (B) illustrate how attention weights work in dialogue act detection task. Figures (C) and (D) represent the key term extraction results. The texts in red represent the ground truth and texts in blue are our prediction results.}}
        \label{fig:vis_att} 
      \end{figure*}

%---------------------------------------------------------------------------------------------%
%---------------------------------------------------------------------------------------------%     
%---------------------------------------------------------------------------------------------%

\subsection{Key Term Extraction} \label{subsec:KE}

%如果空間不夠，可以把以下這段拿掉或縮減
The goal of key term extraction \cite{sarkar2010new,chen2010automatic,nakagawa2002simple,chen2013empirical} is to automatically extract relevant terms from a given document. Key terms may possibly describe the core concept or summary of a document, which can help users understand, organize, and extract important information efficiently from documents. 
These terms are usually manually labeled by humans according to cognition and domain knowledge, so automatic key term extraction is not an easy task. 

Key term extraction can be regarded as a sequence classification problem~\cite{sarkar2010new}.
The model input is a document, while the model selects some terms as key terms from a set of candidates. 
Each term in the set of  candidate terms is considered as a class, and the documents containing the same key terms belong to the same class.
In our task, chances are that some terms do not exist in the document, but they represent the core concepts of the document.
These terms are also regarded as key terms here,  which makes this task even more difficult.
It is possible that a document has more than one key term, or a document can belong to multiple classes.
However, the number of key terms in each testing document is unknown, as a result we consider this task to be a ranking problem. 
That is, the model assigns a score to each candidate term.
Then, the candidate terms are ranked according to the scores.
The target of the system is to rank the key terms above the non key terms.

%In testing procedures, we regarded the predicted vector as a list of relevance scores for each candidate key term. 
%We sorted the relevance scores from high to low and return the sorted list.

In training procedures, each document with $n$ labeled key terms would be mapped into a sparse vector, which is the probability training target. 
The dimension of this sparse vector is the number of candidate terms. 
Most of the values are zero, only the indexes corresponding to labeled terms would be assigned to a value $\frac{1}{n}$, and the summation of this vector is 1. For example, assuming we have 1,000 term candidates and the number of labeled key terms is 4 in a document, we then have an 1,000-dimension sparse target vector with only 4 elements all assigned with $\frac{1}{4}$.

%We applied some modification on it in order to fit the neural network models, which was inspired by Sarkar et.al in \cite{sarkar2010new}. 
%Dialogue act prediction is already a classification problem, but traditional key term extraction task are not. 

  \subsubsection{Experimental setup}
We collected the data from Stack Overflow\footnote{ \url{http://stackoverflow.com/} } website where serves as a platform for users to ask and answer questions. While users of Stack Overflow post questions on the forum, they are asked to label 2\texttildelow 6 key terms for each post. The dataset we collected includes 290,000 examples in total (250,000 for training and 40,000 for testing), and there are about 24,000 kinds of labeled key term. Each example contains a post and 2\texttildelow  6 key term labels, and the average length of the article is about 120 words. The collected dataset is available for download. \footnote{\url{http://speech.ee.ntu.edu.tw/~sense/stackoverflow_pack.zip}}

In practice, to reduce the training complexity, we only selected the 1,000 most frequent key terms in the training set as candidates. These top 1,000 candidates cover over 76\% of the key term labels in the training set, so we can still expect to get reasonable results.

  \subsubsection{Baselines}
  We implemented multiple layer perceptrons (MLP) and long short-term memory (LSTM) networks as the baseline models, which have already been described in section \ref{subsec:baseline}.

\textbf{Tf-idf Sorting} is the baseline we also applied. ``Tf-idf'' is the abbreviation of term frequency-inverse document frequency. It is a numerical statistic that is intended to reflect how important a word is to a document in a collection or corpus. A brief introduction about how Tf-idf Sorting extracts key terms can be found in \cite{Tutorial}.
 We calculated the tf-idf values of a set of candidate key terms according to the dataset, and these candidates were sorted by their values. We then reported the ranking list for evaluation.
  
       \begin{table}[b]
        \caption{\label{tab:ke_scor} {\it The evaluation result for key term extraction.}}
        \vspace{2mm}
        \centerline{
          \begin{tabular}{| l|l | c | c |}
            \hline
            \multicolumn{2}{|l|}{\textbf{Model}} & \multicolumn{1}{c|}{\textbf{MAP (\%)}} & \multicolumn{1}{c|}{\textbf{P@R (\%)}}\\
            \hline \hline
               \multicolumn{2}{|l|}{(a) Oracle} &       \multicolumn{2}{c|}{$77.2$}  \\ \hline
               \multicolumn{2}{|l|}{(b) Tf-idf Sorting} &          $9.9$ & $8.9$ \\  \hline
               \multicolumn{2}{|l|}{(c) Multiple Layer Perceptron} &   $33.1$ & $29.7$ \\ \hline
               \multicolumn{2}{|l|}{(d) Long Short-term Memory} &        $43.1$ & $40.2$ \\ \hline
               \multirow{2}{*}{\footnotesize{Neural Attention Model}} &(e) Sharpening &        $39.3$ & $36.2$ \\ \cline{2-4}
               & (f) Smoothing  &        $\textbf{50.5}$ & $\textbf{46.4}$ \\
            \hline
          \end{tabular}
        }
      \end{table}
      
  \subsubsection{Experimental results}
  
    To examine the prediction result, we chose MAP and P@R as the evaluation methods. The MAP score for a set of documents is the mean of the average precision scores for each document. P@R is defined as the precision after $R$ elements have been selected by the system, where $R$ is also the total number of judged relevant results for the given inputs. Precision is defined as the portion of returned results that are truly belong to the ground truth set.

The experimental results are demonstrated in Table \ref{tab:ke_scor}. Row (a) is the oracle score, which is for reference. Since we only selected 1,000 most frequent key terms as candidates from the training set, we can't achieve 100\% accurate performance. The score of baseline approaches we applied are in rows (b) to (d), and rows (e), (f) are the performance of the proposed neural attention model. The supervised learning baselines outperformed the Tf-idf Sorting baseline (rows (c), (d) vs (b)). That is because without supervised learning, we may not fit the dataset, and we also can't predict the key terms which do not exist in the document. Besides, like the experiment we previously conducted, LSTM shows better ability of handling sequence information in comparison to original neural networks (rows (d) vs (c)), so the LSTM network performs better while using both MAP and P@R as evaluation methods. We found that the performance of our neural attention model with sharpening-attend mechanism degraded while comparing to the original LSTM (rows (e) vs (d), but the one with smoothing attention outperformed all the other approaches (rows (f) vs (b), (c), (d), (e)). This result proved that adding more relevant elements into consideration can help solving sequence classification problems.

%    \begin{equation}  \label{eq:map}
% MAP = \frac{ \sum_{d=1}^{D} Avg P(d) }{D}, 
%\end{equation}

\subsection{Visualization and Analysis} \label{subsec:vis}
Figure \ref{fig:vis_att} demonstrates the visualization of how attention-mechanism works in the sequence classification tasks. The upper row is for dialogue act detection and the lower row is for key term extraction. The darker the color, the higher the weights. We only chose the smoothing-attend mechanism for visualization due to its better performance. According to this figure, we found that attention weights are capable of reducing sentence disfluency problems and filtering out most of the unimportant elements such as function words.

%%%%%%%%%%%%%%%%%%%%%%%%%%%%%%%%%%%%%%%%%%%%%%%%%%%%%
%-------------------------------------------------------------------------------------------------------------------------------%
%%%%%%%%%%%%%%%%%%%%%%%%%%%%%%%%%%%%%%%%%%%%%%%%%%%%%

  \section{Conclusions} \label{sec:conclu}
In this paper, we proposed a neural attention model for sequence classification. In such kinds of task, the input of model is a sequence, and the output is the class of sequence. The major difficulty is that when the input sequence is long, the noisy or irrelevant part may degrade the classification performance. The proposed model can reduce the influences because it is able to highlight important part among the entire sequence. In the experiments, the neural attention model can achieve 72.6\% accuracy for dialogue act detection task and 50.5\% MAP score for key term extraction task, which shows discernible improvements comparing to the other approaches.
% Recurrent neural network architectures combining with attention mechanism, or neural attention model, have shown promising performance recently for the tasks including speech recognition, image caption generation, visual question answering and machine translation.  In this paper, neural attention model is applied on two sequence labeling tasks, dialogue act detection and key term extraction. In the sequence labeling tasks, the model input is a sequence, and the output is the label of the input sequence. The major difficulty of sequence labeling is that when the input sequence is long, it can include many noisy or irrelevant part. If the information in the whole sequence is treated equally, the noisy or irrelevant part may degrade the classification performance.  The attention mechanism is helpful for sequence classification task because it is capable of highlighting important part among the entire sequence for the classification task. The experimental results show that with the attention mechanism, discernible improvements were achieved in the sequence labeling task considered here. The roles of the attention mechanism in the tasks are further analyzed and visualized in this paper.  
\clearpage
 \newpage
  \eightpt
  \bibliographystyle{IEEEtran}

%  \bibliography{mybib,LeeIS16}

\end{document}